%% file: main.tex
\documentclass[10pt,twocolumn,letterpaper]{article}

\usepackage{cvpr}              

\input{preamble}
\definecolor{cvprblue}{rgb}{0.21,0.49,0.74}
\usepackage[pagebackref,breaklinks,colorlinks,allcolors=cvprblue]{hyperref}

\def\paperID{7783} 
\def\confName{CVPR}
\def\confYear{2025}

\title{Skip Tuning: Pre-trained Vision-Language Models are \\ Effective and Efficient Adapters Themselves}

\author{
Shihan Wu\textsuperscript{1} \quad  
Ji Zhang\textsuperscript{2}\thanks{Corresponding author.} \quad  
Pengpeng Zeng\textsuperscript{1} \quad  
Lianli Gao\textsuperscript{1} \quad  
Jingkuan Song\textsuperscript{3}  \quad 
Heng Tao Shen\textsuperscript{3}\\
\textsuperscript{1} University of Electronic Science and Technology of China\\
\textsuperscript{2} Southwest Jiaotong University \\
\textsuperscript{3} Tongji University \\
{\tt\small \textcolor{magenta}{\href{}{\{shihan.wu.koorye@outlook.com,jizhang.jim@gmail.com\}}}}}

\begin{document}
\maketitle

\begin{abstract}
Prompt tuning (PT) has long been recognized as an effective and efficient paradigm for transferring large pre-trained vision-language models (VLMs) to downstream tasks by learning a tiny set of context vectors.
Nevertheless, in this work, we reveal that freezing the parameters of VLMs during learning the context vectors neither facilitates the transferability of pre-trained knowledge nor improves the memory and time efficiency significantly.
Upon further investigation, we find that reducing both the length and width of the feature-gradient propagation flows of the full fine-tuning (FT) baseline is key to achieving effective and efficient knowledge transfer. 
Motivated by this, we propose Skip Tuning, a novel paradigm for adapting VLMs to downstream tasks.
Unlike existing PT or adapter-based methods, Skip Tuning applies Layer-wise Skipping (LSkip) and Class-wise Skipping (CSkip) upon the FT baseline without introducing extra context vectors or adapter modules.
Extensive experiments across a wide spectrum of benchmarks demonstrate the superior effectiveness and efficiency of our Skip Tuning over both PT and adapter-based methods.
Code: \url{https://github.com/Koorye/SkipTuning}.
\end{abstract}

\section{Introduction}

\begin{figure}[t]
\setlength{\abovecaptionskip}{0.1cm}  
\setlength{\belowcaptionskip}{0.cm} 
    \centering
    \includegraphics[width=0.44\textwidth]{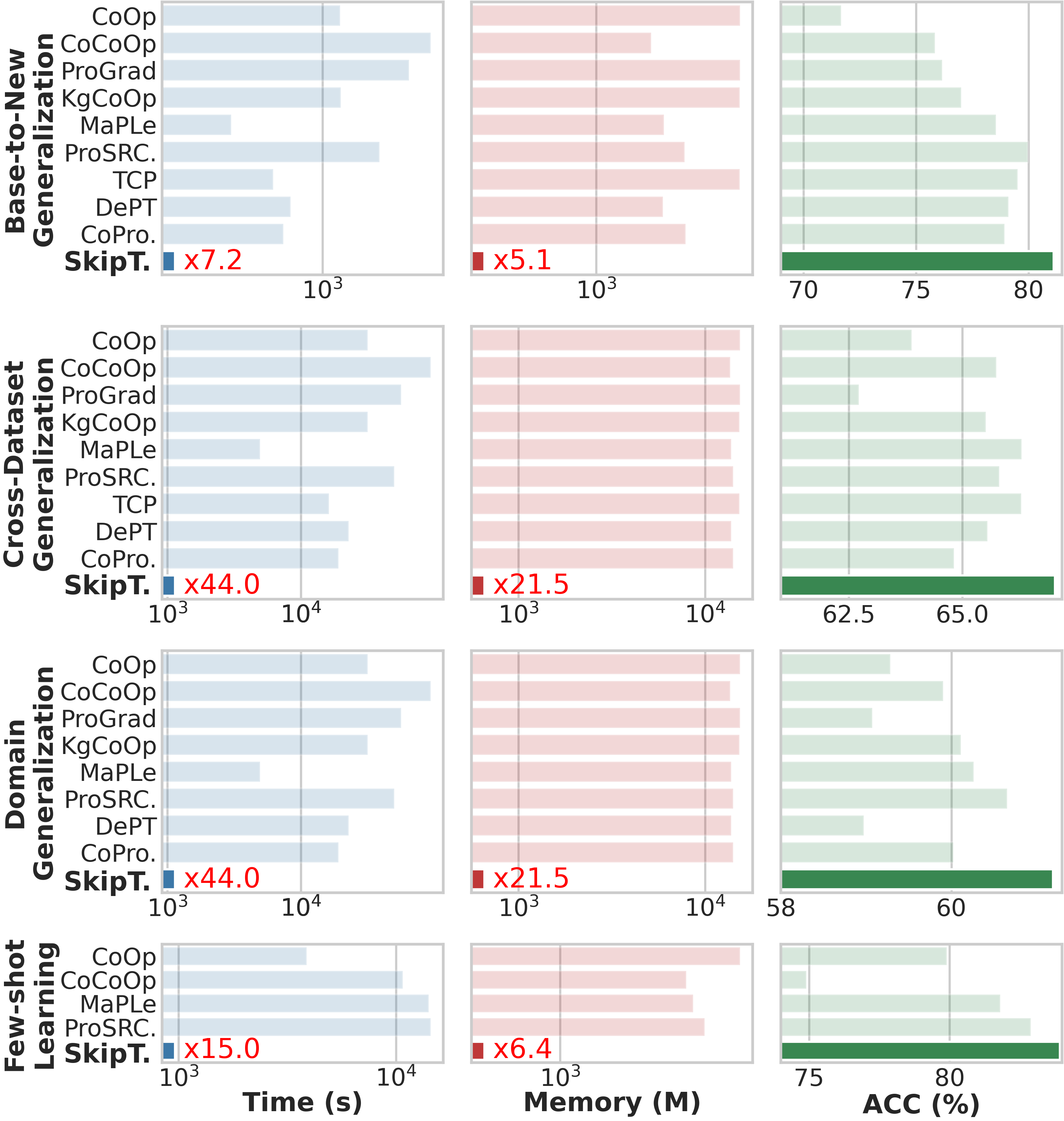}
    \caption{
        Comparison of our devised Skip Tuning with state-of-the-art prompt tuning methods in terms of training time (seconds), memory cost (M), and classification accuracy (\%) across base-to-new generalization, cross-dataset generalization, domain generalization, and few-shot learning benchmarks. 
        {$\times$} indicates the performance improvement over the state-of-the-art.
        Comparison results with the adapter-based methods are reported in Table \ref{tab:adapter}.
    }
    \label{fig:performance}
\end{figure}

\begin{figure}[t]
\setlength{\abovecaptionskip}{0.1cm}  
\setlength{\belowcaptionskip}{0.cm}
    \centering
    \includegraphics[width=0.475\textwidth]{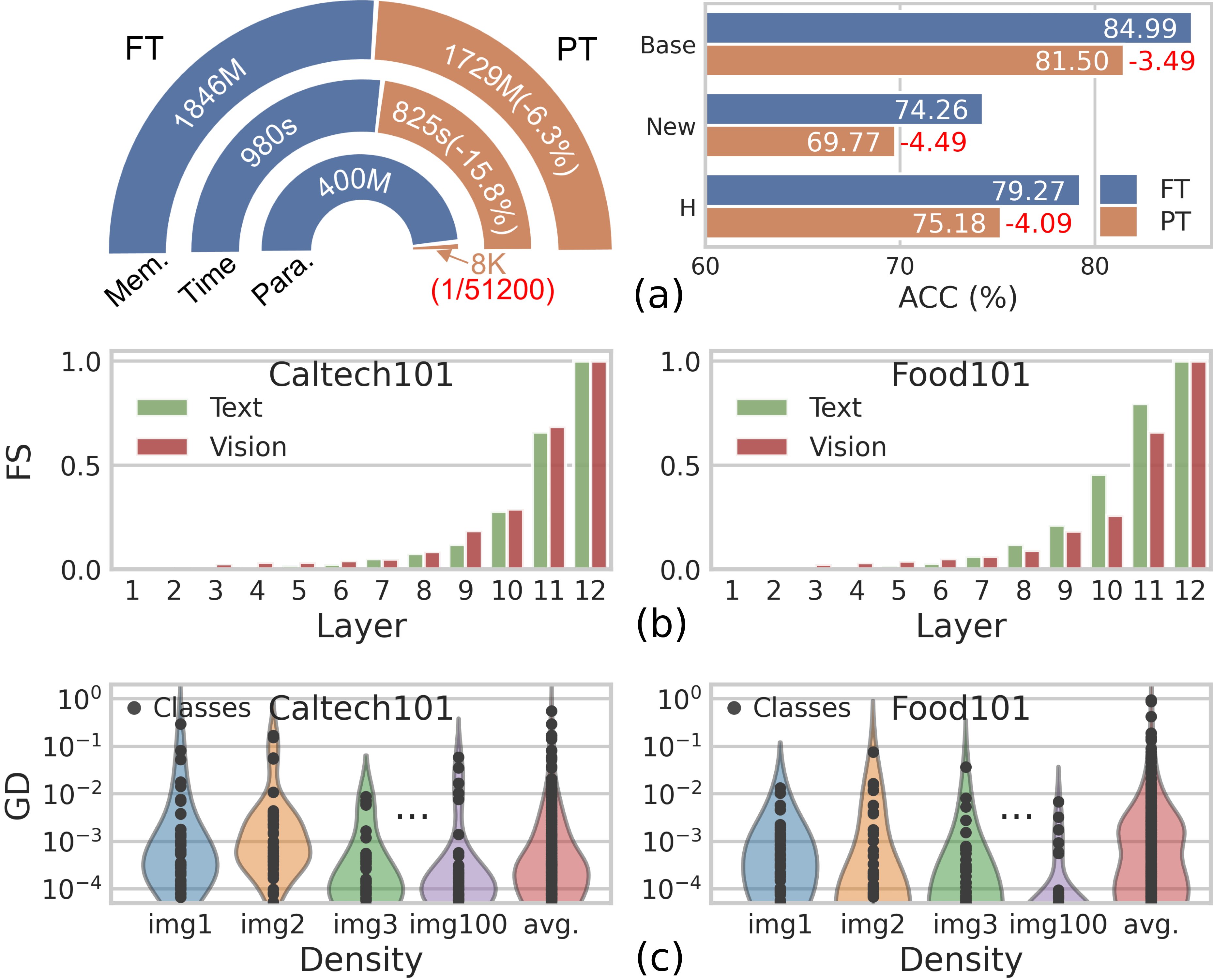}
    \caption{\textbf{Motivations}. \textbf{(a)} Comparison between the prompt tuning (PT) method CoOp \cite{zhou2022coop} and the full fine-tuning (FT) baseline in terms of i) the number of learnable parameters, ii) memory usage, iii) time cost, and iv) base-to-new generalization performance. \textbf{(b)} Feature Sensitivity (\textbf{FS}) of CLIP's network layers, averaged over 100 randomly-sampled training images. \textbf{(c)} Gradient Dependence (\textbf{GD}) of class tokens for different training images.}
    \label{fig:analysis}
\end{figure}

There have recently been significant advancements in large pre-trained vision-language models (VLMs) \cite{radford2021clip,achiam2023gpt,touvron2023llama}. One notable achievement is the CLIP model \cite{radford2021clip}, which leverages an image-text matching (ITM) loss to align images with their corresponding textual descriptions in a common feature space. 
While VLMs have proven impressive capabilities in recognizing open-set visual concepts, their zero-shot generalization performance declines significantly when encountering category, distribution, or domain shifts between upstream training data and downstream tasks.

Prompt tuning (PT) has long been recognized as an effective and efficient paradigm for transferring large pre-trained vision-language models (VLMs) to downstream tasks.
The core concept of PT is to learn a task-specific prompt (i.e., a small number of context vectors) for the target task, using a limited amount of training data, while keeping the pre-trained VLM parameters fixed.
{Although many PT approaches have reported improved performance and parameter efficiency over the full fine-tuning (FT) baseline, the discrepancy of implementation details among those PT approaches obscure the actual performance enhancement.}
For example, the FT performance can be significantly underestimated by training with coarsely tuned hyper-parameters.
Quantitative evidence is presented in Figure \ref{fig:analysis} \textbf{(a)}, where we make comparisons between the PT method CoOp \cite{zhou2022coop} and the FT baseline in terms of the number of learnable parameters,  memory usage, time cost as well as base-to-new generalization performance (see \S \ref{sec:analysis}).
As observed, although PT significantly improves the parameter efficiency of FT (using $\textbf{\underline{1/51200}}$ parameters of FT), the improvements in memory and time efficiency are relatively insignificant (reducing only \textbf{\underline{6.3}}\% memory usage and \textbf{\underline{15.8}}\% time cost).
Besides, compared to FT, the classification accuracy of PT decreases by \textbf{\underline{3.49}}\% and \textbf{\underline{4.49}}\% on base and new tasks, respectively.  
{This suggests that pursuing higher parameter efficiency by freezing the overwhelming majority weights of VLMs during learning the context vectors neither facilitates the transferability of pre-trained knowledge nor improves memory and time efficiency considerably.}
Besides, in many real-world applications, memory and time efficiency often take precedence over parameter efficiency in terms of practical importance.
We therefore raise the following question:

\begin{mdframed}[skipabove=4pt, innertopmargin=4pt, innerbottommargin=4pt]
\textit{Can we optimize the memory and time efficiency of the FT baseline and adapt VLMs to downstream tasks in an effective and efficient manner?}
\end{mdframed}
\vspace{-3.8pt}

\begin{figure*}[t]
\setlength{\abovecaptionskip}{0.2cm}  
\setlength{\belowcaptionskip}{0.cm} 
    \centering
    \includegraphics[width=1\textwidth]{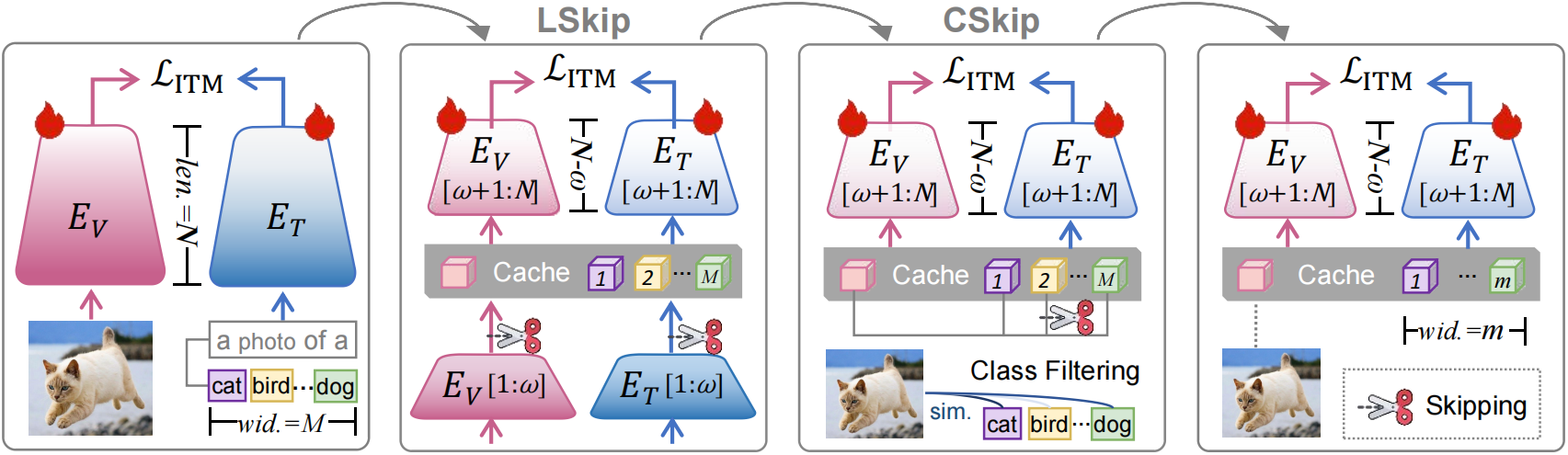}
    \caption{
        \textbf{Overview of our proposed Skip Tuning}.
        Skip Tuning performs Layer-wise Skipping (\textbf{LSkip}) and Class-wise Skipping (\textbf{CSkip})) to enhance the memory and time efficiency of the FT baseline. Specifically,
        LSkip reduces the \textit{length} of feature-gradient propagation flows (FGPFs) by caching intermediate features produced by the $\omega$-th layers of CLIP's vision encoder $E_V$ and text encoder $E_T$ before FT begins. 
        In contrast, CSkip reduces the \textit{width} of FGPFs by filtering out unimportant class tokens in the text encoder $E_T$ for every training image.
    }
    \label{fig:skip_clip}
\end{figure*}

To answer the above question, we scrutinize the Feature-Gradient Propagation Flows (FGPFs) in the vision encoder and text encoder of the CLIP model when performing FT on the base (or target) task. 
Interestingly, we observe that, for each training image, the majority of shallow network layers and class tokens contribute minimally to capturing task-specific knowledge for the base task (see \S \ref{sec:analysis}). 
Motivated by this, we propose Skip Tuning, a novel paradigm for adapting VLMs to downstream tasks without introducing extra context vectors or adapter modules. 
{Concretely, Skip Tuning incorporates two strategies, Layer-wise Skipping (LSkip) and Class-wise Skipping (CSkip), to simultaneously reduce the {length} and {width} of FGPFs in the FT baseline, thereby establishing effective and efficient knowledge transfer of VLMs}, as shown in Figure \ref{fig:skip_clip}.

\noindent\textbf{Effectiveness and Efficiency.} 
We conduct extensive experiments across a wide range of benchmarks to validate the effectiveness and efficiency of our Skip Tuning scheme.
An overview of the achieved results is shown in Figure \ref{fig:performance}.
As seen, our Skip Tuning demonstrates superiority over existing PT methods, e.g., on the few-shot learning benchmark, Skip Tuning achieves  \underline{$\times 
 \textbf{15}$} time efficiency, \underline{$\times \textbf{6.4}$} memory efficiency, while yielding a \underline{\textbf{1.04}}\% improvement in ACC compared to the state-of-the-art.
Furthermore, we also show the advantages of Skip Tuning over existing adapter-based methods in Table \ref{tab:adapter}, where Skip Tuning achieves \underline{$\times \textbf{3.8}$} time efficiency, \underline{$\times \textbf{3.9}$} memory efficiency, along with a \underline{\textbf{3.59}}\% H ACC enhancement over the strong rival LoRA \cite{hu2021lora}.

\noindent\textbf{Main Contributions.} 
The main contributions are threefold. 
\begin{itemize}
\item We reveal that reducing both the width and length of the feature-gradient propagation flows (FGPFs) of the full fine-tuning (FT) baseline is key to establishing effective and efficient knowledge transfer.
\item We devise Skip Tuning, an effective and efficient method for transferring VLMs to downstream tasks without relying on extra context vectors or adapter modules.
\item We evaluate our method on a wide spectrum of benchmarks, demonstrating the superiority of Skip Tuning over both prompt tuning and adapter-based approaches.
\end{itemize}

\section{Methodology}

In this section, we elaborate on our Skip Tuning approach.
We start with an introduction of preliminaries. 

\subsection{Preliminaries}

\noindent\textbf{Contrastive Language-Image Pre-training (CLIP).} 
Following the common practice of existing paradigms for adapting PVLMs, we adopt CLIP \cite{radford2021clip} as the testbed in this work.
CLIP aims to learn an alignment between image and text features generated by an image encoder and a text encoder, respectively. By being exposed to 400 million image-text association pairs and employing a contrastive learning paradigm within a shared feature space, CLIP acquires a diverse array of open-set visual concepts that can be readily applied to downstream tasks. For instance, zero-shot classification can be achieved by framing the classification task as an image-text matching problem. Initially, a prompt (e.g. “\texttt{a photo of a}”) is crafted to extract text features of all intra-task classes by presenting the class-extended prompt (“\texttt{a photo of a [CLS]}”) to the text encoder. Subsequently, the image encoder is utilized to derive the image feature of a given input example, and class prediction is performed by comparing cosine distances between the image feature and text features of classes.

\noindent\textbf{Prompt Tuning with an Image-Text Matching Loss.} 
PT aims to learn a task-specific prompt with few-labeled samples from the base task. 
Concretely, a prompt is formulated as ${\kappa}$ context vectors: $\{\boldsymbol{v}_1,\boldsymbol{v}_2,...,\boldsymbol{v}_{\kappa}\}$. 
During training, we produce the text feature of the $j$-th class by inputting the class-token-extended prompt $\boldsymbol{{c}}_j=\{\boldsymbol{v}_1,\boldsymbol{v}_2,...,\boldsymbol{v}_{\kappa},[\texttt{CLS}]\}$ to the text encoder $g(\cdot)$, where $[\texttt{CLS}]$ indicates the class token of the $j$-th class.
Denote $\boldsymbol{f}_i \in \mathbb{R}^{d}$ as the image feature of a sample $\boldsymbol{x}_i$ extracted by the image encoder, the task-specific prompt can be updated via back-propagating the image-text matching (ITM) loss through the frozen CLIP model. The ITM loss can be expressed as
\begin{equation}
    \mathcal{L}_{\mathtt{ITM}}=-\sum_j \boldsymbol{y}_j \log \mathrm{p}(\boldsymbol{{c}}_j| \boldsymbol{x}_i),
    \label{eq.1}
\end{equation}
where $\boldsymbol{y}$ is the one-hot label, and
\begin{equation}
    \mathrm{p}(\boldsymbol{{c}}_j | \boldsymbol{x}_i)=\frac{e^{<g(\boldsymbol{{c}}_j), \boldsymbol{f}_i>/\tau}}{\sum_{t=1}^{M}e^{<g(\boldsymbol{{c}}_t), \boldsymbol{f}_i>/\tau}},
    \label{inf}
\end{equation}
 $<\cdot>$ denotes cosine similarity, $M$ is the number of classes, and $\tau$ is the temperature learned by CLIP.

\subsection{Rethinking the FT Baseline}
\label{sec:analysis}
While many PT schemes have reported improved performance and efficiency over the full fine-tuning (FT) baseline, the discrepancy of the implementation details among those schemes obscures the actual performance enhancement.  

\noindent\textbf{Comparison of PT and FT.}
To comprehensively evaluate the progress established by PT, we perform a comparison between the representative PT method CoOp \cite{zhou2022coop} with the full FT baseline in terms of i) the number of learnable parameters, ii) memory usage, iii) time cost, and iv) base-to-new generalization results. 
The experimental setting is shown in \textbf{Sup. Mat. (A)}. 
From the obtained results in Figure \ref{fig:analysis} \textbf{(a)}, we observe that although PT significantly reduces the number of learnable parameters compared with FT (using $\textbf{\underline{1/51200}}$ parameters of FT), the improvements in memory efficiency and time efficiency are relatively insignificant (reducing only \textbf{\underline{6.3}}\% memory usage and \textbf{\underline{15.8}}\% time cost).
Notably, in comparison to FT, the classification accuracy of PT decreases by \textbf{\underline{3.49}}\% and \textbf{\underline{4.49}}\% on base and new tasks, respectively.
{This means that pursuing higher parameter efficiency by freezing the vast majority of VLM weights during PT neither enhances the transferability of pre-trained knowledge nor improves memory and time efficiency significantly.}
In other words, existing PT methods involve a trade-off between parameter efficiency and performance.

\noindent\textbf{Cost Analysis of FT.} 
The computational cost (i.e., memory and time cost) for fine-tuning VLMs is mainly caused by Feature-Gradient Propagation Flows (FGPFs)—the forward propagation flows of image/text features and the backward propagation flows of gradients. 
For example, on the CLIP model, each propagation flow is required to traverse all the $N$ layers of the vision encoder $E_V$ and the text encoder $E_T$.
For simplicity, we assume that $E_V$ and $E_T$ have identical network structures (e.g. ViT-B/16).
Denote the number of class tokens as $M$, and the costs of the FGPFs in the $l$-th layer of $E_V$ and $E_T$ as $C_V$ and $C_T$ respectively. 
Thus, the total cost $C_{\rm{total}}=N\times (C_V + C_T\times M\footnote{For each training image, we need to obtain the text features of all $M$ classes to calculate the image-text matching loss, as illustrated in Eq. (\ref{inf}).})$. 
In this work, we refer to the number of network layers and the number of class tokens \textit{used for each training image} as the \textit{length} and \textit{width} of FGPFs, respectively. 
In particular, $\textit{length}=N$, $\textit{width}=M$ for existing PT methods, as shown in Figure \ref{fig:skip_clip}.
We therefore raise the following question:
\begin{mdframed}[skipabove=4pt, innertopmargin=4pt, innerbottommargin=4pt]
\textit{Can we reduce both the length and width of FGPFs without compromising the FT performance?}
\end{mdframed}
\vspace{-1pt}

\noindent\textbf{Towards Effective and Efficient Knowledge Transfer.} 
To answer the above question, we design two metrics of Feature Sensitivity (\textbf{FS}) and Gradient Dependence (\textbf{GD}) to estimate the contributions of different network layers and class tokens for each training image $\boldsymbol{x}$, respectively.

More specifically, let $\boldsymbol{f}_l$ and ${\boldsymbol{f}_l^{\prime}}$ be a vision (or a text) feature before and after fine-tuning the $l$-th layer of the CLIP's vision (or text) encoder, the \textbf{FS} of the $l$-th layer for the training image $\boldsymbol{x}$ can be expressed as
\begin{equation}
    \mathbf{FS}_l(\boldsymbol{x})=\psi(\boldsymbol{f}_l,{\boldsymbol{f}_l^{\prime}}),
\end{equation}
where $\psi(\cdot)$ is the Euclidean distance. Hence, the bigger the $\mathbf{FS}_l(\boldsymbol{x})$ value, the more important the $l$-th layer for the input image $\boldsymbol{x}$.
Let $\nabla{\boldsymbol{f}_l}$ indicate the computed feature gradients passing through the last layer of the vision encoder for the training image $\boldsymbol{x}$ and $\nabla{\boldsymbol{f}_l^{(c)}}$ be the feature gradients after the $c$-th class token is removed from Eq. \ref{inf}. The \textbf{GD} of the $c$-th class for $\boldsymbol{x}$ can be computed as
\begin{equation}
    \mathbf{GD}_c (\boldsymbol{x}) = \psi(\nabla{\boldsymbol{f}_l}, \nabla{\boldsymbol{f}_l^{(c)}}).
\end{equation}
Therefore, the bigger the $\mathbf{GD}_c (\boldsymbol{x})$ value, the more important the $c$-th class for the input image $\boldsymbol{x}$.

Figure \ref{fig:analysis} \textbf{(b)} illustrates the obtained \textbf{FS} values of different image/text encoder layers when adapting CLIP to the two datasets Caltech101 and Food101. 
As can be observed, the FS scores of the vast majority of shallow layers are close to 0, and only the last few layers contribute significantly to adapting CLIP to the two datasets.
Moreover, Figure \ref{fig:analysis} \textbf{(c)} presents the frequency distributions of the \textbf{GD} values of class tokens for 100 training images randomly sampled from the two datasets. 
As shown, most class tokens contribute minimally to capturing task-specific knowledge from each training image.
In a nutshell, the above observations reveal that we can achieve effective and efficient FT by filtering out unimportant network layers and class tokens in the text encoder for every training image. 

\subsection{Skip Tuning}
\label{sec:skip_tuning}
Motivated by the previous section, we propose Skip Tuning, which performs Layer-wise Skipping (\textbf{LSkip}) and Class-wise Skipping (\textbf{CSkip}) to adapt VLMs to downstream tasks in an effective and efficient manner, as shown in Figure \ref{fig:skip_clip}.

\noindent\textbf{Layer-wise Skipping (\textbf{LSkip}).}
LSkip aims to reduce the length of FGPFs without compromising the performance of the FT baseline.
To this end, LSkip first saves the image and text features before the $\omega$-th shallow layers of $E_V$ and $E_T$ in a cache, and then uses the saved intermediate features as input to update the parameters of the remaining $N-\omega$ deep layers. 
Denote $\boldsymbol{v}_i^{\omega}$ and $\boldsymbol{t}_j^{\omega}$ as the extracted image and text features of the image $\boldsymbol{x}_i$ and the class token $\texttt{CLS}$ ($j=1,...,{M}$) in the $\omega$-th layers of $E_V$ and $E_T$, respectively, i.e.,
\begin{equation}
    \boldsymbol{v}_i^{\omega}=E_V[1:\omega](\boldsymbol{x}_i), \boldsymbol{t}_j^{\omega}=E_T[1:\omega](\boldsymbol{c}_j),
\end{equation}
where $\boldsymbol{c}_j$ is constructed using a hand-crafted prompt, e.g.,
$\boldsymbol{c}_j=\texttt{a photo of a [CLS]}$.

After that, the $[1:\omega]$ layers of both $E_V$ and $E_T$ are discarded.
During FT, we input $\boldsymbol{v}_i^{\omega}$ and $\boldsymbol{t}_j^{\omega}$ to the remaining $N-\omega$ deep layers of $E_V$ and $E_T$, and obtain:
\begin{equation}
    \tilde{\boldsymbol{v}}_i^{N}=E_V[\omega+1: N](\boldsymbol{v}_i^{\omega}), \tilde{\boldsymbol{t}}_j^{N}=E_T[\omega+1: N](\boldsymbol{t}_j^{\omega}),
\end{equation}
which are used to calculate the loss $\mathcal{L}_{\mathrm{ITM}}$ in Eq. \ref{eq.1}.
In this way, the image and text features, along with the calculated gradients, propagate through only $N-\omega$ layers, effectively reducing the length of FGPFs.

\noindent\textbf{Class-wise Skipping (\textbf{CSkip}).}
Given a training image, the goal of CSkip is to filter out unimportant class tokens in the text encoder $E_T$ during the construction of the image-text matching loss $\mathcal{L}_{\mathrm{ITM}}$ (Eq. \ref{eq.1}).
Intuitively, the direct way is to select the top $k$ closest class tokens to compute the loss for each image, based on the similarities between the image feature and the text features of the $M$ inner-task classes. 
Nevertheless, we empirically find that this strategy leads to overfitting by selecting the same subset of classes for each image across different training epochs. 
Therefore, we propose an exponential image-conditioned class filtering strategy to overcome this limitation.

Concretely, before FT begins, we use the text encoder $E_T$ to produce \textit{M} class features with a hand-crafted prompt, e.g. “\texttt{a photo of a [CLS]}”. 
Then, we sort the cosine similarities between the vision feature of a training image and the \textit{M} class features in descending order. 
The probability of sampling the $j$-th class token for the current training image can be expressed as:
\begin{equation}
    p_j=\begin{cases}
        1 &  o_j \le r\times M,\\
        \zeta(o_j-r\times M,\lambda) & o_j>r\times M, \\
    \end{cases}
\end{equation}
where $\zeta(\iota,\lambda)=e^{-\lambda \iota}$, $\lambda>0$ is the exponential decay coefficient, 
$r$ is the sampling ratio, and $o_j$ is the sorting index of the $j$-th class.
In this way, for each image, we can maintain the top $r\times M$ closest class tokens while also sampling the remaining $(1-r) \times M$ classes with a certain probability for constructing $\mathcal{L}_{\mathrm{ITM}}$, improving the diversity of training data at different epochs.
Denote $m$ as the number of class tokens sampled for the $i$-th training image, we have $m \ll M$.
This means we can flexibly reduce the width of FGPFs in the text encoder for every training image. 
Surprisingly, our experimental results in the next section reveal that ignoring the majority of class tokens for each training image can enhance the generalization performance of the learned model. 
This improvement may stem from CSkip’s capability to filter out redundant and distracting text features when performing image-text matching with the loss of $\mathcal{L}_{\mathrm{ITM}}$.

\section{Experiments}
In this section, we demonstrate the efficiency and effectiveness of our devised Skip Tuning scheme on base-to-new generalization, cross-dataset generalization, domain generalization, and few-shot learning benchmarks.

\begin{table*}[h]
\setlength{\abovecaptionskip}{0.cm}  
\setlength{\belowcaptionskip}{-0.2cm} 
    \renewcommand\arraystretch{0.8}
    \setlength{\tabcolsep}{5.5pt}
    \footnotesize
    \centering
    \caption{Base-to-new generalization results over 11 datasets. * indicates our reproduced results.}
    \label{tab:base_to_new}
    \tabcolsep 0.071in
    \begin{tabular}{ll|ccccccccc|c}
        \specialrule{1pt}{0.5pt}{2pt}
        \multicolumn{1}{l}{\multirow{2}{*}{{{Datasets}}}}& \multicolumn{1}{l|}{\multirow{2}{*}{{{Metric}}}} & CoOp & CoCoOp & ProGrad & KgCoOp & MaPLe & ProSRC. & TCP & DePT & CoPro.* & \textbf{SkipT.} \\
        & & \scriptsize{(IJCV'22)} & \scriptsize{(CVPR'22)} & \scriptsize{(ICCV'23)} & \scriptsize{(CVPR'23)} & \scriptsize{(CVPR'23)} & \scriptsize{(ICCV'23)} & \scriptsize{(CVPR'24)} & \scriptsize{(CVPR'24)} & \scriptsize{(ICLR'24)} & \scriptsize{(Ours)} \\
        \hline
        \multirow{3}{*}{ImgNet} & Base & 76.47 & 75.98 & 77.02 & 75.83 & 76.66 & 77.60 & 77.27 & 77.03 & 76.53 & \textbf{77.73} \\
        & New & 67.88 & 70.43 & 66.66 & 69.96 & 70.54 & 70.73 & 69.87 & 70.13 & \textbf{71.30} & 70.40 \\
        & H & 71.92 & 73.10 & 71.46 & 72.78 & 73.47 & \textbf{74.01} & 73.38 & 73.42 & 73.82 & 73.89 \\
        \hline
        \multirow{3}{*}{Caltech} & Base & 98.00 & 97.96 & 98.02 & 97.72 & 97.74 & 98.10 & 98.23 & 98.30 & \textbf{98.60} & 98.50 \\
        & New & 89.31 & 93.81 & 93.89 & 94.39 & 94.36 & 94.03 & 94.67 & 94.60 & 95.17 & \textbf{95.33} \\
        & H & 93.73 & 95.84 & 95.91 & 96.03 & 96.02 & 96.02 & 96.42 & 96.41 & 96.85 & \textbf{96.89} \\
        \hline
        \multirow{3}{*}{Pets} & Base & 93.67 & 95.20 & 95.07 & 94.65 & 95.43 & 95.33 & 94.67 & 94.33 & 94.73 & \textbf{95.70} \\
        & New & 95.29 & 97.69 & 97.63 & 97.76 & 97.76 & 97.30 & 97.20 & 97.23 & 96.70 & \textbf{97.87} \\
        & H & 94.47 & 96.43 & 96.33 & 96.18 & 96.58 & 96.30 & 95.92 & 95.76 & 95.71 & \textbf{96.77} \\
        \hline
        \multirow{3}{*}{Cars} & Base & 78.12 & 70.49 & 77.68 & 71.76 & 72.94 & 78.27 & 80.80 & 79.13 & 73.17 & \textbf{82.93} \\
        & New & 60.40 & 73.59 & 68.63 & 75.04 & 74.00 & 74.97 & 74.13 & \textbf{75.47} & 70.63 & 72.50 \\
        & H & 68.13 & 72.01 & 72.88 & 73.36 & 73.47 & 76.58 & 77.32 & 77.26 & 71.88 & \textbf{77.37} \\
        \hline
        \multirow{3}{*}{Flowers} & Base & 97.60 & 94.87 & 95.54 & 95.00 & 95.92 & 98.07 & 97.73 & 98.00 & 96.93 & \textbf{98.57} \\
        & New & 59.67 & 71.75 & 71.87 & 74.73 & 72.46 & \textbf{76.50} & 75.57 & 76.37 & 75.50 & 75.80 \\
        & H & 74.06 & 81.71 & 82.03 & 83.65 & 82.56 & \textbf{85.95} & 85.23 & 85.84 & 84.88 & 85.70 \\
        \hline
        \multirow{3}{*}{Food101} & Base & 88.33 & 90.70 & 90.37 & 90.50 & \textbf{90.71} & 90.67 & 90.57 & 90.50 & 90.37 & 90.67 \\
        & New & 82.26 & 91.29 & 89.59 & 91.70 & \textbf{92.05} & 91.53 & 91.37 & 91.60 & 91.53 & 92.03 \\
        & H & 85.19 & 90.99 & 89.98 & 91.09 & \textbf{91.38} & 91.10 & 90.97 & 91.05 & 90.95 & 91.34 \\
        \hline
        \multirow{3}{*}{Aircraft} & Base & 40.44 & 33.41 & 40.54 & 36.21 & 37.44 & 42.73 & 41.97 & 43.20 & 36.17 & \textbf{45.37} \\
        & New & 22.30 & 23.71 & 27.57 & 33.55 & 35.61 & \textbf{37.87} & 34.43 & 34.83 & 34.47 & 37.13 \\
        & H & 28.75 & 27.74 & 32.82 & 34.83 & 36.50 & 40.15 & 37.83 & 38.57 & 35.30 & \textbf{40.84} \\
        \hline
        \multirow{3}{*}{SUN397} & Base & 80.60 & 79.74 & 81.26 & 80.29 & 80.82 & \textbf{82.67} & 82.63 & 82.33 & 82.30 & 82.40 \\
        & New & 65.89 & 76.86 & 74.17 & 76.53 & 78.70 & 78.47 & 78.20 & 77.80 & \textbf{79.63} & 79.03 \\
        & H & 72.51 & 78.57 & 77.55 & 78.36 & 79.75 & 80.52 & 80.35 & 80.00 & \textbf{80.94} & 80.68 \\
        \hline
        \multirow{3}{*}{DTD} & Base & 79.44 & 77.01 & 77.35 & 77.55 & 80.36 & 83.37 & 82.77 & 82.20 & 83.00 & \textbf{83.77} \\
        & New & 41.18 & 56.00 & 52.35 & 54.99 & 59.18 & 62.97 & 58.07 & 59.13 & 63.20 & \textbf{67.23} \\
        & H & 54.24 & 64.85 & 62.45 & 64.35 & 68.16 & 71.15 & 68.25 & 68.78 & 71.76 & \textbf{74.59} \\
        \hline
        \multirow{3}{*}{EuroSAT} & Base & 92.19 & 87.49 & 90.11 & 85.64 & \textbf{94.07} & 92.90 & 91.63 & 89.03 & 93.77 & 92.47 \\
        & New & 54.74 & 60.04 & 60.89 & 64.34 & 73.23 & 73.90 & 74.73 & 71.07 & 71.73 & \textbf{83.00} \\
        & H & 68.69 & 71.21 & 72.67 & 73.48 & 82.35 & 82.32 & 82.32 & 79.04 & 81.28 & \textbf{87.48} \\
        \hline
        \multirow{3}{*}{UCF101} & Base & 84.69 & 82.33 & 84.33 & 82.89 & 83.00 & 87.10 & 87.13 & 85.80 & 86.20 & \textbf{87.30} \\
        & New & 56.05 & 73.45 & 74.94 & 76.67 & 78.66 & 78.80 & 80.77 & 77.23 & 78.70 & \textbf{82.47} \\
        & H & 67.46 & 77.64 & 79.35 & 79.65 & 80.77 & 82.74 & 83.83 & 81.29 & 82.28 & \textbf{84.81} \\
        \hline
        \rowcolor{green!6} & Base & 82.69 & 80.47 & 82.48 & 80.73 & 82.28 & 84.26 & 84.13 & 83.62 & 82.89 & \textbf{85.04} \\
        \rowcolor{green!6} & New & 63.22 & 71.69 & 70.75 & 73.60 & 75.14 & 76.10 & 75.36 & 75.04 & 75.32 & \textbf{77.53} \\
        \rowcolor{green!6} \multirow{-3}{*}{\textbf{Avg ACC}} & H & 71.66 & 75.83 & 76.16 & 77.00 & 78.55 & 79.97 & 79.51 & 79.10 & 78.93 & \textbf{81.11} \\
        \hline        
         \rowcolor{red!5}& Time (s)& 1186 & 2851 & 2311 & 1191 & 413 & 1735 & 619 & 733 & 684 & \textbf{239} \\
        \rowcolor{red!5} \multirow{-2}{*}{\textbf{Cost}} & Mem. (M) & 3204 & 1556 & 3204 & 3188 & 1729 & 2041 & 3189 & 1714 & 2060 & \textbf{404} \\
        \specialrule{1pt}{0.5pt}{2pt}
    \end{tabular}
\end{table*}
\begin{table*}[h]
\setlength{\abovecaptionskip}{0.cm}  
\setlength{\belowcaptionskip}{-0.2cm} 
    \renewcommand\arraystretch{0.8}
    \setlength{\tabcolsep}{5.5pt}
    \footnotesize
    \centering
    \caption{Cross-dataset generalization results on 11 datasets. * indicates our reproduced results. The detailed results on the 10 target datasets (i.e, Caltech, Pets, Cars, ... , and UCF101) are reported in \textbf{Sup. Mat. (B)}.}
    \label{tab:cross_dataset}
    \tabcolsep 0.069in
    \begin{tabular}{ll|ccccccccc|c}
        \specialrule{1pt}{0.5pt}{2pt}
        \multicolumn{1}{l}{\multirow{2}{*}{{{Datasets}}}}& \multicolumn{1}{l|}{\multirow{2}{*}{{{Metric}}}} & CoOp & CoCoOp & ProGrad & KgCoOp & MaPLe & ProSRC. & TCP & DePT & CoPro.* & \textbf{SkipT.} \\
        & & \scriptsize{(IJCV'22)} & \scriptsize{(CVPR'22)} & \scriptsize{(ICCV'23)} & \scriptsize{(CVPR'23)} & \scriptsize{(CVPR'23)} & \scriptsize{(ICCV'23)} & \scriptsize{(CVPR'24)} & \scriptsize{(CVPR'24)} & \scriptsize{(ICLR'24)} & \scriptsize{(Ours)} \\
        \specialrule{.4pt}{0.5pt}{0.5pt}
        \rowcolor{green!6} ImageNet & ACC   & 71.51 & 71.02 & 72.24 & 70.66 & 70.72 & 71.27 & 71.40 & \textbf{72.77} & 72.53 & \textbf{72.77} \\
        \rowcolor{green!6} 10 Datasets & Avg ACC  & 63.88 & 65.74 & 62.71 & 65.51 & 66.30 & 65.81 & 66.29 & 65.55 & 64.81 & \textbf{67.00} \\
        \specialrule{.4pt}{0.5pt}{0.5pt}
          \rowcolor{red!5} & Time (s) & 31632 & 93917 & 56223 & 31636 & 4942 & 50091 & 16174 & 22796 & 19161 & \textbf{1139} \\
        \rowcolor{red!5} \multirow{-2}{*}{\textbf{Cost}} & Mem. (M) & 15412 & 13622 & 15412 & 15254 & 13786 & 14107 & 15263 & 13783 & 14131 & \textbf{656} \\
        \specialrule{1pt}{0.5pt}{2pt}
    \end{tabular}
\end{table*}
\begin{table*}[!h]
\setlength{\abovecaptionskip}{0.cm}  
\setlength{\belowcaptionskip}{-0.2cm} 
    \renewcommand\arraystretch{1}
    \setlength{\tabcolsep}{9.5pt}
    \footnotesize
    \centering
    \caption{Domain generalization results on ImageNet. * indicates our reproduced results. The detailed results on the 4 ImgNet variants (i.e.,  ImgNet-V2, ImgNet-S, ImgNet-A, and  ImgNet-R) are presented in \textbf{Sup. Mat. (B)}.}
    \label{tab:domain_generalization}
    \tabcolsep 0.087in
    \begin{tabular}{ll|cccccccc|c}
        \specialrule{1pt}{2pt}{0.5pt}
        \multicolumn{1}{l}{\multirow{2}{*}{{{Datasets}}}}& \multicolumn{1}{l|}{\multirow{2}{*}{{{Metric}}}}  & CoOp & CoCoOp & ProGrad & KgCoOp & MaPLe & ProSRC. & DePT & CoPro.* & \textbf{SkipT.} \\
        & & \scriptsize{(IJCV'22)} & \scriptsize{(CVPR'22)} & \scriptsize{(ICCV'23)} & \scriptsize{(CVPR'23)} & \scriptsize{(CVPR'23)} & \scriptsize{(ICCV'23)} &  \scriptsize{(CVPR'24)}& \scriptsize{(ICLR'24)} & \scriptsize{(Ours)} \\
        \hline
        \rowcolor{green!6} ImageNet & ACC  & 71.51 & 71.02 & 72.24 & 70.66 & 70.72 & 71.27 & \textbf{72.77} & 72.53 & \textbf{72.77} \\
        \rowcolor{green!6} 4 Imag.Variants & Avg ACC  & 59.28 & 59.90 & 59.07 & 60.11 & 60.26 & 60.65 & 58.97 & 60.02 & \textbf{61.20} \\
        \specialrule{.4pt}{0.5pt}{0.5pt}
        \rowcolor{red!5}  & Time (s) & 31632 & 93917 & 56223 & 31636 & 4942 & 50091 & 22796 & 19161 & \textbf{1139} \\
        \rowcolor{red!5}  \multirow{-2}{*}{\textbf{Cost}} & Mem. (M) & 15412 & 13622 & 15412 & 15254 & 13786 & 14107 & 13783 & 14131& \textbf{656} \\
        \specialrule{1pt}{0.5pt}{2pt}
    \end{tabular}
\end{table*}

\subsection{Experimental Setup}

\noindent\textbf{Datasets.} 
We conduct experiments using 11 datasets from diverse sources. In particular, for base-to-new generalization, cross-dataset transfer and few-shot learning, we use 11 datasets including ImageNet \cite{deng2009imagenet}, Caltech101 \cite{fei2007caltech}, OxfordPets \cite{parkhi2012pets}, StanfordCars \cite{krause2013cars}, Flowers102 \cite{nilsback2008flowers}, Food101 \cite{bossard2014food}, FGVCAircraft \cite{maji2013aircraft}, SUN397 \cite{xiao2010sun}, UCF101 \cite{soomro2012ucf101}, DTD \cite{cimpoi2014dtd} and EuroSAT \cite{helber2019eurosat}. 
For domain generalization, we use ImageNet \cite{deng2009imagenet} as the source dataset and its four variants as target datasets including ImageNetV2 \cite{recht2019imagenetv2}, ImageNet-Sketch \cite{wang2019sketch}, ImageNet-A \cite{hendrycks2021imageneta}, and ImageNet-R \cite{hendrycks2021imagenetr}. 

\noindent\textbf{Evaluation Metric.}
We report base-task accuracy (\%, denoted as \textit{Base}), new-task accuracy (\%, denoted as \textit{New}), and their harmonic mean (\%, denoted as \textit{H}) to compare the performance/effectiveness of different methods.
We also report the time cost (seconds, denoted as \textit{Time}) and memory usage (M, denoted as \textit{Memory}) for efficiency evaluation.

\noindent\textbf{Implementation details.} 
Our implementation of Skip Tuning is based on the open-source Github repository of DePT \cite{zhang2024dept}\footnote{\url{https://github.com/Koorye/DePT}}.
In concrete terms, we leverage pre-trained ViT-B/16 as the backbone of the CLIP model. We employ the SGD optimizer to train the model with a learning rate of 2e-5 and a batch size of 4.
By default, the number of training epochs is set to 20 for the base-to-new generalization, cross-dataset generalization, and domain generalization benchmarks, and 40 for the few-shot learning benchmark.
The above hyper-parameters are fixed across all datasets.
We adjust the LSkip hyper-parameter $\omega$, CSkip hyper-parameters $r$, and $\lambda$ in $\S$ \ref{section.abl}. 
All experimental results are the average of 3 runs with different seeds.
We conduct experiments using an {NVIDIA V100} GPU. 
For more details and additional results, please refer to \textbf{Sup. Mat.}

\subsection{Experimental Results} 
\noindent\textbf{Base-to-New Generalization.} 
The {base-to-new generalization} setting evaluates whether the models learned on a base task can generalize to new tasks with unseen classes. 
Following the comparison methods, for each dataset, we first construct a base task and a new task by equally dividing the dataset into two sets of classes, then we perform prompt tuning on the base task and test the learned model on both the base and new tasks.
Table \ref{tab:base_to_new} presents the obtained results of Skip Tuning and other state-of-the-art prompt tuning methods on 11 datasets.
As shown, Skip Tuning achieves the best base-to-new generalization performance with the lowest memory usage and time cost.
Concretely, Skip Tuning achieves \underline{$\times \textbf{7.2}$} time efficiency, and \underline{$\times \textbf{5.1}$} memory efficiency, while maintaining a \underline{\textbf{1.14}}\% improvement in H ACC compared to the previous state-of-the-art method PromptSRC. This demonstrates the effectiveness and efficiency of our proposed Skip Tuning method.
Moreover, we observe a tradeoff between base-task and new-task accuracies for most competitors. For instance, CoPrompt outperforms DePT on new tasks but lags behind DePT on base tasks. Notably, our Skip Tuning achieves the best performance on both base and new tasks simultaneously. This suggests that Skip Tuning effectively mitigates the overfitting issue when transferring VLMs to the base (or target) task.

\noindent\textbf{Cross-Dataset Generalization.}
The cross-dataset generalization setting assesses whether models trained on a source dataset/distribution can generalize to unseen target datasets/distributions.
We follow the common setup of those comparison methods to use ImageNet as the source dataset and the other 10 datasets as target datasets. 
Table \ref{tab:cross_dataset} presents the obtained cross-dataset generalization results of our Skip Tuning and other state-of-the-art methods on the source and target datasets.
As seen, compared to the previous state-of-the-art method PromptSRC, our Skip Tuning achieves \underline{$\times \textbf{44}$} time efficiency, and \underline{$\times \textbf{21.5}$} memory efficiency, while establishing \underline{\textbf{1.5}}\% and  \underline{\textbf{1.19}}\% ACC improvements on the source and target distributions respectively. 
From the average results, Skip Tuning consistently shows superior effectiveness and efficiency over the nine competitors on the 10 target datasets, without compromising the results of the tuned model on the source dataset. 
This demonstrates the effectiveness of our Skip Tuning scheme for improving the robustness of the tuned model to distribution shifts.

\noindent\textbf{Domain Generalization.}
The domain generalization setting assesses whether models trained on a source domain can generalize to unseen/target domains.
In line with those comparison methods, we consider the ImageNet as the source domain and the other four ImageNet variants as target domains.
As shown in Table \ref{tab:cross_dataset}, compared to the previous state-of-the-art method PromptSRC, Skip Tuning achieves \underline{$\times \textbf{44}$} time efficiency, and \underline{$\times \textbf{21.5}$} memory efficiency, while establishing \underline{\textbf{1.5}}\% and  \underline{\textbf{0.55}}\% ACC improvements on the source and target domains, respectively.
Besides, Skip Tuning consistently outperforms those competitors on the four target domains without compromising the performance of the tuned model on the source domain, which proves the effectiveness of our Skip Tuning scheme in improving the robustness of the tuned model to domain shifts. 

\subsection{Ablation Studies}
\label{section.abl}

\begin{figure*}[t]
\setlength{\abovecaptionskip}{0.1cm}  
\setlength{\belowcaptionskip}{0.cm} 
    \centering
    \includegraphics[width=0.93\textwidth]{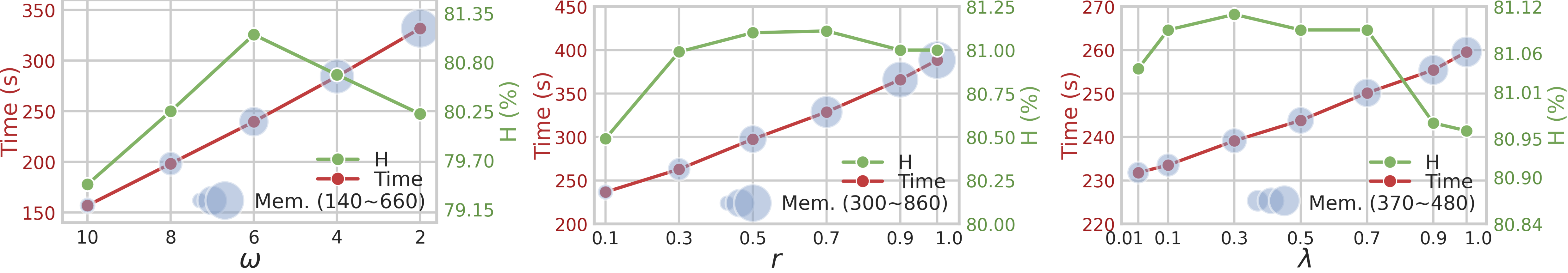}
    \caption{Ablation studies of the number of skipped layers $\omega$ in LSkip, and the sampling rate $r$, the decay coefficient $\lambda$ in CSkip.}
    \label{fig:ablation}
\end{figure*}

In this section, we conduct ablative studies to further scrutinize our devised Skip Tuning method. 

\noindent\textbf{Effectiveness of the Designed Components.}
Our proposed Skip Tuning approach simultaneously performs Layer-wise Skipping (LSkip) and Class-wise Skipping (CSkip) to enhance the memory and time efficiency of the full fine-tuning (FT) baseline.
In this experiment, we investigate the effectiveness of LSkip and CSkip by gradually adding them to the FT baseline.
From the results in Table \ref{tab:ablation}, we have the following observations.
\textbf{{{i)}}} Both LSkip and CSkip contribute to performance improvement.
\textbf{{{ii)}}} By combining all those components, Skip Tuning improves the effectiveness and efficiency of the baseline method remarkably. 
\textbf{{{iii)}}} The memory and time efficiency gains are more pronounced when CSkip is applied to the FT baseline, compared to the performance improvements it yields.
\textbf{{{iv)}}} CSkip improves the memory and time efficiency of the baseline without compromising the generalization performance of the learned model. One possible reason is that CSkip can filter out redundant and distracting text features when performing image-text matching with the loss of $\mathcal{L}_{\mathrm{ITM}}$.

\begin{table}[t]
\setlength{\abovecaptionskip}{0.cm}  
\setlength{\belowcaptionskip}{-0.2cm} 
    \renewcommand\arraystretch{0.8}
    \setlength{\tabcolsep}{5pt}
    \footnotesize
    \centering
    \caption{Ablation study on the components of Skip Tuning.}
    \label{tab:ablation}
    \tabcolsep 0.09in
    \begin{tabular}{cc|ccc|cc}
        \specialrule{1pt}{2pt}{0.5pt}
        LSP& CSP& Base & New & H & Time (s) & Mem. (M) \\
        \specialrule{.4pt}{0.5pt}{0.5pt}
        \scriptsize{\XSolidBrush} & \scriptsize{\XSolidBrush} & 84.99 & 74.26 & 79.27 & 1002 & 1846 \\
        \specialrule{.4pt}{0.5pt}{0.5pt}
         \scriptsize{\Checkmark}& \scriptsize{\XSolidBrush} & \textbf{85.08} & 77.30 & 81.00 & 388 & 861 \\
        \specialrule{.4pt}{0.5pt}{0.5pt}
         \scriptsize{\XSolidBrush} & \scriptsize{\Checkmark} & 84.77 & 74.48 & 79.29 & 440 & 919 \\
        \specialrule{.4pt}{0.5pt}{0.5pt}
         \scriptsize{\Checkmark}& \scriptsize{\Checkmark}& \cellcolor{green!6}85.04 & \cellcolor{green!6}\textbf{77.53}& \cellcolor{green!6}\textbf{81.11} &\cellcolor{red!5}\textbf{239}  & \cellcolor{red!5}\textbf{404} \\
        \specialrule{1pt}{0.5pt}{2pt}
    \end{tabular}
\end{table}
\begin{table}[t]
\setlength{\abovecaptionskip}{-0.cm}  
\setlength{\belowcaptionskip}{-0.2cm} 
    \renewcommand\arraystretch{0.8}
    \setlength{\tabcolsep}{7.5pt}
    \footnotesize
    \centering
    \caption{Comparison of our Skip Tuning and adapter-based methods. 
    $^{\dagger}$ denotes the efficiency-optimized versions (details are illustrated in \textbf{Sup. Mat. (C)}).}
    \label{tab:adapter}
    \tabcolsep 0.076in
    \begin{tabular}{l|ccc|cc}
        \specialrule{1pt}{2pt}{0.5pt} 
        Method & Base & New & H & Time (s) & Mem. (M) \\
        \hline
        FT & 84.99 & 74.26 & 79.27 & 1002 & 1846 \\
        CLIP-adapter \cite{gao2024clipadapter} & 74.48 & 73.81 & 74.14 & 888 & 1784 \\
        LoRA \cite{hu2021lora} & 80.53 & 74.73 & 77.52 & 910 & 1580 \\
        \textbf{SkipT.} (Ours) & \cellcolor{green!6}\textbf{85.04} & \cellcolor{green!6}\textbf{77.53} & \cellcolor{green!6}\textbf{81.11} & \cellcolor{red!5}\textbf{239} & \cellcolor{red!5}\textbf{404} \\
        \hline
        CLIP-adapter$^{\dagger}$ & 76.75 & 73.56 & 75.12 & 137 & 175 \\
        \textbf{SkipT.} (Ours)$^{\dagger}$ & \cellcolor{green!6}\textbf{82.66} &\cellcolor{green!6}\textbf{75.56} & \cellcolor{green!6}\textbf{78.95} & \cellcolor{red!5}\textbf{115} & \cellcolor{red!5}\textbf{67} \\
        \specialrule{1pt}{0.5pt}{2pt}
    \end{tabular}
\end{table}

\noindent\textbf{Impact of the LSkip Hyper-parameter $\omega$.}
Our Skip Tuning approach drops out the $1\sim \omega$ layers of the CLIP's vision and text encoders in the LSkip step. 
Here, we investigate the impact of $\omega$ on performance by setting $\omega$ to the values of $\{2,4,6,8,10\}$. 
The obtained average testing results on the 11 datasets are reported in Figure \ref{fig:ablation} ({\textit{Left}}).
As shown, the obtained H ACCs gradually increase as the $\omega$ value decreases from 10 to 6, after which the performance gradually decreases. 
But, we also see that as the value of $\omega$ becomes smaller, both memory and time costs increase.
Hence, we set $\omega=6$  for LSKip in this work.

\noindent\textbf{Impact of the CSkip Hyper-parameter $r$.}
In the CSkip step of Skip Tuning, we devise an image-conditioned class sampling strategy to filter out irrelevant class tokens for each training image. 
The larger the sampling ratio $r$ value, the more class tokens will be used to construct the training loss $\mathcal{L}_{\mathrm{ITM}}$ for each image.
It is necessary to scrutinize the impact of $r$ on performance.
To this end, we respectively set $r$ to $\{0.1, 0.3, 0.5, 0.7, 0.9, 1.0\}$, and report the average testing results on the 11 datasets in Figure \ref{fig:ablation}  ({\textit{Mid.}}).
From the obtained H ACCs, our method is in general not sensitive to the change of  $r$ within a certain range (from 0.4 to 1.0).
We also see a gradual decrease in H ACC when $r>0.7$, suggesting that for each training image, not all class tokens are beneficial for capturing task-specific knowledge. 
Besides, we can observe that as the value of $r$ becomes larger, both memory and time costs increase.
Therefore, we set $r=0.5$  for CSKip in this work.

\noindent\textbf{Impact of the CSkip Hyper-parameter $\lambda$.}
Our Skip Tuning introduces an exponential decay coefficient $\lambda$ for the devised image-conditioned class sampling strategy in the CSkip step. 
We study the impact of $\lambda$ by setting $\lambda$ to the values of $\{0.01, 0.1, 0.3, 0.5, 0.7, 0.9, 1.0\}$, and report the average testing results over the 11 datasets in Figure \ref{fig:ablation} ({\textit{Right}}).
As can be observed from the obtained H ACCs, our method is in general not sensitive to $\lambda$ when it takes the values from 0.1 to 0.7.
Particularly, the H ACCs gradually decrease as $\lambda$ increases from 0.7 to 1.0.
Also, we can see that as the value of $\lambda$ becomes larger, both memory and time costs increase.
Thus, we set $\lambda=0.3$  for CSkip in this work.

\begin{figure*}[t]
\setlength{\abovecaptionskip}{0.2cm}  
\setlength{\belowcaptionskip}{-0.3cm} 
    \centering
    \includegraphics[width=0.95\textwidth]{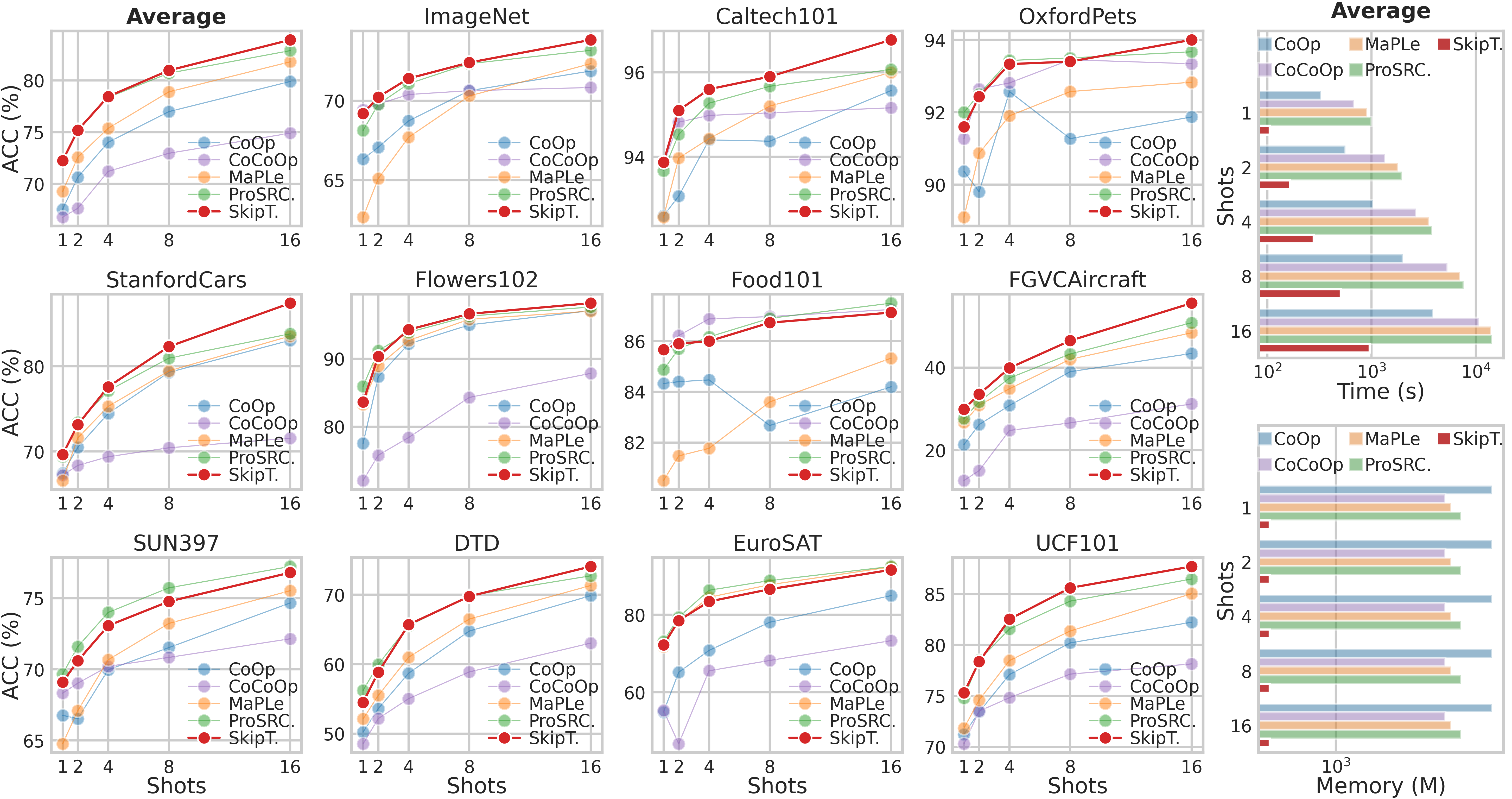}
    \caption{Few-shot learning results on 11 datasets. For detailed results, please visit \textbf{Sup. Mat. (D)}}
    \label{fig:few_shot}
\end{figure*}

\subsection{Additional Results}
\noindent\textbf{Comparison with Adapter-based Methods.}
Previous experimental results demonstrate the superior effectiveness and efficiency of our Skip Tuning method over existing prompt tuning methods.
To further demonstrate the advantages of Skip Tuning, we also compare it with the representative adapter-based methods LoRA \cite{hu2021lora} and CLIP-adapter \cite{gao2024clipadapter}. The base-to-new generalization results averaged on 11 datasets are reported in Table \ref{tab:adapter}, where
CLIP-adapter$^{\dagger}$ refers to our re-implementation of CLIP-adapter, i.e., the CLIP's last-layer features are cached once before training and subsequently used to update the weights of the adapter module. As observed, our Skip Tuning offers substantial memory and time efficiency advantages over the two strong competitors, while also delivering better classification performance on both base and new tasks.

\noindent\textbf{Few-shot Learning.}
In the previous experiments, we follow the comparison approaches to evaluate the performance of different methods on \textit{M}-way 16-shot tasks—16 training examples are sampled for each of the \textit{M} inner-task classes.
It is interesting to further scrutinize the effectiveness and efficiency of our Skip Tuning method under different shots.
To this end, Figure \ref{fig:few_shot} reports the few-shot learning performance of Skip Tuning and other representative competitors.
As shown, our Skip Tuning demonstrates superior performance compared to other methods, with significant efficiency advantages in memory usage and time cost across 1-shot to 16-shot settings. 
Additionally, while PromptSRC and MaPLe enhance CoOp's performance, both methods incur increased time cost. 
In contrast, Skip Tuning achieves both effectiveness and efficiency without compromise, further underscoring the superiority of our approach.

\section{Related Work}
\noindent\textbf{Pre-trained Vision-Language Models.} 
Pre-trained vision-language models (VLMs) have garnered great attention recently.
Notable models such as CLIP\cite{radford2021clip}, ALIGN\cite{jia2021align}, LiT\cite{zhai2022lit} and FILIP\cite{yao2021filip} have demonstrated exceptional performance across various vision-language tasks.
These models are pre-trained on extensive datasets comprising image-text pairs sourced from the internet.
For instance, CLIP\cite{radford2021clip} and ALIGN\cite{jia2021align} respectively utilize over 400 million and 1 billion pairs,  for training.
The large scale of training data enables these models to excel in open-vocabulary image-text retrieval and zero-shot classification.
Additionally, they have been successfully applied to downstream tasks including image classification\cite{gao2024clipadapter,zhang2021tip}, object detection\cite{gu2021vild,feng2022promptdet,fu2023styleadv,zang2022ovdetr,zhou2022detic}, and semantic segmentation\cite{luddecke2022clipseg,rao2022denseclip}. 
In this work, we focus on the effective adaptation of vision-language models to downstream visual recognition tasks.

\noindent\textbf{Prompt Tuning.} 
Prompt tuning (a.k.a. context optimization \cite{zhou2022coop}) has emerged as a parameter-efficient learning paradigm to adapt powerful VLMs to downstream tasks \cite{zhang2023deta,zhang2022free,fu2025cross,fu2022generalized}. 
As a representative method, CoOp \cite{zhou2022coop} enables task adaptation by optimizing a set of prompt vectors within CLIP’s language branch. 
While effective, CoOp often suffers from limited generalization on new tasks due to overfitting to the base task. 
A series of schemes are devised to tackle this problem, e.g. CoCoOp\cite{zhou2022cocoop}, 
KgCoOp\cite{yao2023kgcoop}, ProGrad\cite{zhu2023prograd}, TCP \cite{yao2024tcp}, and DePT\cite{zhang2024dept}.
By adding trainable prompts into both the image and text branches of CLIP, MaPLe\cite{khattak2023maple}, PromptSRC\cite{khattak2023promptsrc}, and CoPrompt\cite{roy2023coprompt} demonstrate remarkable performance on both base and new tasks.
Despite the advantages, we reveal that freezing the parameters of VLMs during learning the context vectors neither facilitates the generalization of pre-trained knowledge nor significantly improves memory and time efficiency.

\section{Conclusion}
In this work, we first reveal that freezing the parameters of VLMs during prompt tuning neither facilitates the transferability of pre-trained knowledge
nor improves memory and time efficiency considerably.
To circumvent this limitation, we propose Skip Tuning, an effective and efficient method for transferring VLMs to downstream tasks without relying on extra context vectors or adapter modules.
Extensive experiments across a broad range of benchmarks demonstrate the superiority of our Skip Tuning method over both prompt tuning and adapter-based approaches.

\noindent\textbf{Acknowledgements.} 
This study is supported by grants from the National Natural Science Foundation of China (Grant No. 62425208, No. 62122018, No. 62020106008), the Postdoctoral Fellowship Program of CPSF (Grant No. GZB20240625), the Science and Technology Innovation Committee of Shenzhen Municipality Foundation (Grant No.JCYJ20240813114208012), and Kuaishou Tech.

{
    \small
    \bibliographystyle{ieeenat_fullname}
    \bibliography{main}
}

\newpage
\appendix

\noindent\textbf{\Large{\, \, \, \, \, \, Supplementary Materials}}

\section{Experimental Setups for FT and PT}
While many Prompt Tuning (PT) schemes have reported improved performance and efficiency over the full fine-tuning (FT) baseline, the discrepancy of the implementation details among those schemes obscures the actual performance enhancement.  
To comprehensively evaluate the progress established by PT, in Section 2.2 of the paper, we perform a comparison between the representative PT method CoOp \cite{zhou2022coop} with the FT baseline in terms of i) the number of learnable parameters, ii) memory usage, iii) time cost, and iv) base-to-new generalization results. The experimental setups of FT and PT are as follows.

\begin{itemize}
    \item For PT method CoOp, we use the publicly available ViT-B/16 CLIP model\cite{radford2021clip} as feature backbone.
Following the experimental setting used by Muhammad et al.\cite{khattak2023maple}, we employ the SGD optimizer with a learning rate of 3.5e-3, a batch size of 4, and training apochs of 10.
\item  The FT baseline involves updating all parameters of the pre-trained CLIP model. For fair comparisons, we maintain the same batch size of 4 and a total of 10 training epochs (as CoOp), the learning rate is set to 1e-5.
\end{itemize}

\section{Cross-Dataset / Domain General. Results}

\begin{table*}[h]
\setlength{\abovecaptionskip}{0.cm}  
\setlength{\belowcaptionskip}{-0.2cm} 
    \renewcommand\arraystretch{0.8}
    \setlength{\tabcolsep}{5.5pt}
    \footnotesize
    \centering
    \caption{Cross-dataset generalization results on 11 datasets. * indicates our reproduced results. }
    \label{tab:cross_dataset}
    \tabcolsep 0.067in
    \begin{tabular}{ll|cccccccccc}
        \specialrule{1pt}{2pt}{0.5pt}
        & & CoOp & CoCoOp & ProGrad & KgCoOp & MaPLe & ProSRC. & TCP & DePT & CoPro.* & SkipT. \\
        & & \scriptsize{(IJCV'22)} & \scriptsize{(CVPR'22)} & \scriptsize{(ICCV'23)} & \scriptsize{(CVPR'23)} & \scriptsize{(CVPR'23)} & \scriptsize{(ICCV'23)} & \scriptsize{(CVPR'24)} & \scriptsize{(CVPR'24)} & \scriptsize{(ICLR'24)} & \scriptsize{(Ours)} \\
        \specialrule{.4pt}{0.5pt}{0.5pt}
        \rowcolor{green!5} Source & ImgNet & 71.51 & 71.02 & 72.24 & 70.66 & 70.72 & 71.27 & 71.40 & \textbf{72.77} & 72.53 & \textbf{72.77} \\
        \specialrule{.4pt}{0.5pt}{0.5pt}
        \rowcolor{green!5} & \textbf{Average} & 63.88 & 65.74 & 62.71 & 65.51 & 66.30 & 65.81 & 66.29 & 65.55 & 64.81 & \textbf{67.00} \\
        \multirow{9}{*}{Target} & Caltech & 93.70 & \textbf{94.43} & 91.52 & 93.92 & 93.53 & 93.60 & 93.97 & 94.23 & 94.37 & 93.43 \\
        & Pets & 89.14 & 90.14 & 89.64 & 89.83 & 90.49 & 90.25 & \textbf{91.25} & 90.03 & 89.50 & 90.10 \\
        & Cars & 64.51 & 65.32 & 62.39 & 65.41 & 65.57 & \textbf{65.70} & 64.69 & 65.57 & 65.57 & 65.37 \\
        & Flowers & 68.71 & 71.88 & 67.87 & 70.01 & \textbf{72.23} & 70.25 & 71.21 & 70.57 & 69.63 & 71.97 \\
        & Food101 & 85.30 & 86.06 & 85.40 & 86.36 & 86.20 & 86.15 & \textbf{86.69} & 86.37 & 85.03 & 86.17 \\
        & Aircraft & 18.47 & 22.94 & 20.16 & 22.51 & 24.74 & 23.90 & 23.45 & 23.27 & 23.47 & \textbf{25.13} \\
        & SUN397 & 64.15 & \textbf{67.36} & 62.47 & 66.16 & 67.01 & 67.10 & 67.15 & 66.67 & 67.13 & 67.33 \\
        & DTD & 41.92 & 45.73 & 39.42 & 46.35 & 46.49 & 46.87 & 44.35 & 44.97 & 44.33 & \textbf{48.00} \\
        & EuroSAT & 46.39 & 45.37 & 43.46 & 46.04 & 48.06 & 45.50 & 51.45 & 43.53 & 40.87 & \textbf{54.27} \\
        & UCF101 & 66.55 & 68.21 & 64.29 & 68.50 & 68.69 & 68.75 & 68.73 & \textbf{69.30} & 68.17 & 68.23 \\
        \specialrule{.4pt}{0.5pt}{0.5pt}
        \rowcolor{red!6} & \textbf{Time (s)} & 31632 & 93917 & 56223 & 31636 & 4942 & 50091 & 16174 & 22796 & 19161 & \textbf{1139} \\
        \rowcolor{red!6} \multirow{-2}{*}{\textbf{Cost}} & \textbf{Mem. (M)} & 15412 & 13622 & 15412 & 15254 & 13786 & 14107 & 15263 & 13783 & 14131 & \textbf{656} \\
        \specialrule{1pt}{0.5pt}{2pt}
    \end{tabular}
\end{table*}
\begin{table*}[!h]
\setlength{\abovecaptionskip}{0.cm}  
\setlength{\belowcaptionskip}{-0.2cm} 
    \renewcommand\arraystretch{1}
    \setlength{\tabcolsep}{9.5pt}
    \footnotesize
    \centering
    \caption{Domain generalization results on ImageNet. * indicates our reproduced results.}
    \label{tab:domain_generalization}
    \tabcolsep 0.092in
    \begin{tabular}{ll|ccccccccc}
        \specialrule{1pt}{2pt}{0.5pt}
        & & CoOp & CoCoOp & ProGrad & KgCoOp & MaPLe & ProSRC. & DePT & CoPro.* & SkipT. \\
        & & \scriptsize{(IJCV'22)} & \scriptsize{(CVPR'22)} & \scriptsize{(ICCV'23)} & \scriptsize{(CVPR'23)} & \scriptsize{(CVPR'23)} & \scriptsize{(ICCV'23)} & \scriptsize{(CVPR'24)} & \scriptsize{(ICLR'24)} & \scriptsize{(Ours)} \\
        \specialrule{.4pt}{0.5pt}{0.5pt}
        \rowcolor{green!5} Source & ImgNet & 71.51 & 71.02 & 72.24 & 70.66 & 70.72 & 71.27 & \textbf{72.77} & 72.53 & \textbf{72.77} \\
        \specialrule{.4pt}{0.5pt}{0.5pt}
        \rowcolor{green!5} & \textbf{Average} & 59.28 & 59.90 & 59.07 & 60.11 & 60.26 & 60.65 & 58.97 & 60.02 & \textbf{61.20} \\
        \multirow{3}{*}{Target} & ImgNet-V2 & 64.20 & 64.07 & 64.73 & 64.10 & 64.07 & 64.35 & 64.70 & 65.40 & \textbf{65.67} \\
        & ImgNet-S & 47.99 & 48.75 & 47.61 & 48.97 & 49.15 & 49.55 & 47.73 & \textbf{49.90} & 49.73 \\
        & ImgNet-A & 49.71 & 50.63 & 49.39 & 50.69 & 50.90 & 50.90 & 48.33 & 47.53 & \textbf{51.13} \\
        & ImgNet-R & 75.21 & 76.18 & 74.58 & 76.70 & 76.98 & 77.80 & 75.10 & 77.23 & \textbf{78.27} \\
        \specialrule{.4pt}{0.5pt}{0.5pt}
        \rowcolor{red!6} & \textbf{Time (s)} & 31632 & 93917 & 56223 & 31636 & 4942 & 50091 & 22796 & 19161 & \textbf{1374} \\
        \rowcolor{red!6} \multirow{-2}{*}{\textbf{Cost}} & \textbf{Mem. (M)} & 15412 & 13622M & 15412 & 15254 & 13786 & 14107 & 13783 & 14131 & \textbf{925} \\
        \specialrule{1pt}{0.5pt}{2pt}
    \end{tabular}
\end{table*}

To evaluate the robustness of different methods to distribution and domain shifts, we conduce comprehensive comparisons in cross-dataset generalization and domain generalization benchmarks.

\textbf{Cross-Dataset Generalization.} 
The cross-dataset generalization setting assesses whether models trained on a source dataset/distribution can generalize to unseen target datasets/distributions.
We follow the common setup of those comparison methods to use ImageNet as the source dataset and the other 10 datasets as target datasets. Table \ref{tab:cross_dataset} presents the obtained cross-dataset generalization results of our Skip Tuning and other state-of-the-art methods on the source and target datasets.
As shown, Skip Tuning achieves the highest overall performance while maintaining the lowest time and memory costs, especially on challenging datasets with significant differences from the source domain (e.g., Aircraft, DTD, EuroSAT).
This demonstrates the effectiveness of our Skip Tuning scheme for improving the robustness of the tuned model to distribution shifts. 

\textbf{Domain Generalization.}
The domain generalization setting assesses whether models trained on a source domain can generalize to unseen/target domains.
In line with those comparison approaches, we consider the ImageNet dataset as the source domain and the other four ImageNet variants as target domains.
Table \ref{tab:domain_generalization} reports the obtained domain generalization results of our Skip Tuning and other state-of-the-art methods on the source and target ddomains.
As can be seen, our Skip Tuning outperforms other methods in terms of overall performance and maintains the lowest time and memory costs.
Besides, Skip Tuning consistently outperforms those competitors on the four target domains without compromising the performance of the tuned model on the source domain, which proves the effectiveness of  our Skip Tuning scheme in improving the robustness of the tuned model to domain shifts. 

\section{Comparisons with Adapter-based Methods}


\begin{table}[h]
\setlength{\abovecaptionskip}{0.cm}  
\setlength{\belowcaptionskip}{-0.2cm} 
    \small
    \centering
    \caption{Details adapter-based methods and our Skip Tuning.}
    \label{tab:adapter}
    \begin{tabular}{c|cccc|c}
        \specialrule{1pt}{2pt}{0.5pt}
        Method & Cache & LR & BS & Epochs & $\omega$ \\
        \specialrule{.4pt}{0.5pt}{0.5pt}
        LoRA & - & 1e-3 & 4 & 10 & - \\
        CLIP-adapter & \XSolidBrush & 2e-3 & 32 & 100 & - \\
        CLIP-adapter$^{\dagger}$ & \Checkmark & 3.5e-3 & 4 & 10 & - \\
        \hline
        SkipT.(Ours) & \Checkmark & 2e-5 & 4 & 10 & 6 \\
        SkipT.(Ours)$^{\dagger}$ & \Checkmark & 2e-5 & 4 & 10 & 10 \\
        \specialrule{1pt}{0.5pt}{2pt}
    \end{tabular}
\end{table}

\begin{table*}[t]
\setlength{\abovecaptionskip}{0.cm}  
\setlength{\belowcaptionskip}{-0.2cm} 
    \footnotesize
    \centering
    \caption{Few-shot Learning results on 11 datasets.}
    \label{tab:few_shot}
    \begin{tabular}{lc|cccccc}
        \specialrule{1pt}{2pt}{0.5pt}
        \multirow{2}{*}{Shot} & \multirow{2}{*}{Dataset} & CoOp & CoCoOp & MaPLe & ProSRC. & SkipT. \\
        & & (IJCV'22) & (CVPR'22) & (CVPR'23) & (ICCV'23) & (Ours) \\
        \specialrule{.4pt}{0.5pt}{0.5pt}
         & ImgNet & 66.33 & \textbf{69.43} & 62.67 & 68.13 & 69.20 \\
         & Caltech & 92.60 & 93.83 & 92.57 & 93.67 & \textbf{93.87} \\
         & Pets & 90.37 & 91.27 & 89.10 & \textbf{92.00} & 91.60 \\
         & Cars & 67.43 & 67.22 & 66.60 & 69.40 & \textbf{69.63} \\
         & Flowers & 77.53 & 72.08 & 83.30 & \textbf{85.93} & 83.63 \\
         & Food101 & 84.33 & 85.65 & 80.50 & 84.87 & \textbf{85.67} \\
         & Aircraft & 21.37 & 12.68 & 26.73 & 27.67 & \textbf{29.93} \\
         & SUN397 & 66.77 & 68.33 & 64.77 & \textbf{69.67} & 69.10 \\
         & DTD & 50.23 & 48.54 & 52.13 & \textbf{56.23} & 54.50 \\
         & EuroSAT & 54.93 & 55.33 & 71.80 & \textbf{73.13} & 72.23 \\
         & UCF101 & 71.23 & 70.30 & 71.83 & 74.80 & \textbf{75.30} \\
        \rowcolor{green!6} & \textbf{Average} & 67.56 & 66.79 & 69.27 & \textbf{72.32} & 72.24 \\
        \rowcolor{red!5} & \textbf{Time (s)} & 327 & 678 & 911 & 999 & \textbf{106} \\
        \rowcolor{red!5}\multirow{-14}{*}{1} & \textbf{Mem. (M)} & 4551 & 2888 & 3061 & 3373 & \textbf{528} \\
        \specialrule{.4pt}{0.5pt}{0.5pt}
         & ImgNet & 67.07 & 69.78 & 65.10 & 69.77 & \textbf{70.23} \\
         & Caltech & 93.07 & 94.82 & 93.97 & 94.53 & \textbf{95.10} \\
         & Pets & 89.80 & \textbf{92.64} & 90.87 & 92.50 & 92.43 \\
         & Cars & 70.50 & 68.37 & 71.60 & \textbf{73.40} & 73.17 \\
         & Flowers & 87.33 & 75.79 & 88.93 & \textbf{91.17} & 90.33 \\
         & Food101 & 84.40 & \textbf{86.22} & 81.47 & 85.70 & 85.90 \\
         & Aircraft & 26.20 & 15.06 & 30.90 & 31.70 & \textbf{33.53} \\
         & SUN397 & 66.53 & 69.03 & 67.10 & \textbf{71.60} & 70.60 \\
         & DTD & 53.60 & 52.17 & 55.50 & \textbf{59.97} & 58.83 \\
         & EuroSAT & 65.17 & 46.74 & 78.30 & \textbf{79.37} & 78.50 \\
         & UCF101 & 73.43 & 73.51 & 74.60 & \textbf{78.50} & 78.37 \\
        \rowcolor{green!6} & \textbf{Average} & 70.65 & 67.65 & 72.58 & \textbf{75.29} & 75.18 \\
        \rowcolor{red!5} & \textbf{Time (s)} & 561 & 1346 & 1785 & 1947 & \textbf{165} \\
        \rowcolor{red!5}\multirow{-14}{*}{2} & \textbf{Mem. (M)} & 4551 & 2888 & 3061 & 3373 & \textbf{528} \\
        \specialrule{.4pt}{0.5pt}{0.5pt}
         & ImgNet & 68.73 & 70.39 & 67.70 & 71.07 & \textbf{71.40} \\
         & Caltech & 94.40 & 94.98 & 94.43 & 95.27 & \textbf{95.60} \\
         & Pets & 92.57 & 92.81 & 91.90 & \textbf{93.43} & 93.33 \\
         & Cars & 74.47 & 69.39 & 75.30 & 77.13 & \textbf{77.60} \\
         & Flowers & 92.17 & 78.40 & 92.67 & 93.87 & \textbf{94.27} \\
         & Food101 & 84.47 & \textbf{86.88} & 81.77 & 86.17 & 86.00 \\
         & Aircraft & 30.83 & 24.79 & 34.87 & 37.47 & \textbf{39.90} \\
         & SUN397 & 69.97 & 70.21 & 70.67 & \textbf{74.00} & 73.07 \\
         & DTD & 58.70 & 55.04 & 61.00 & 65.53 & \textbf{65.70} \\
         & EuroSAT & 70.80 & 65.56 & 84.50 & \textbf{86.30} & 83.40 \\
         & UCF101 & 77.10 & 74.82 & 78.47 & 81.57 & \textbf{82.53} \\
        \rowcolor{green!6} & \textbf{Average} & 74.02 & 71.21 & 75.37 & 78.35 & \textbf{78.44} \\
        \rowcolor{red!5} & \textbf{Time (s)} & 1036 & 2684 & 3541 & 3841 & \textbf{280} \\
        \rowcolor{red!5}\multirow{-14}{*}{4} & \textbf{Mem. (M)} & 4551 & 2888 & 3061 & 3373 & \textbf{528} \\
        \specialrule{.4pt}{0.5pt}{0.5pt}
         & ImgNet & 70.63 & 70.63 & 70.30 & 72.33 & \textbf{72.40} \\
         & Caltech & 94.37 & 95.04 & 95.20 & 95.67 & \textbf{95.90} \\
         & Pets & 91.27 & 93.45 & 92.57 & \textbf{93.50} & 93.40 \\
         & Cars & 79.30 & 70.44 & 79.47 & 80.97 & \textbf{82.33} \\
         & Flowers & 94.97 & 84.30 & 95.80 & 96.27 & \textbf{96.60} \\
         & Food101 & 82.67 & \textbf{86.97} & 83.60 & 86.90 & 86.73 \\
         & Aircraft & 39.00 & 26.61 & 42.00 & 43.27 & \textbf{46.50} \\
         & SUN397 & 71.53 & 70.84 & 73.23 & \textbf{75.73} & 74.77 \\
         & DTD & 64.77 & 58.89 & 66.50 & \textbf{69.87} & 69.77 \\
         & EuroSAT & 78.07 & 68.21 & 87.73 & \textbf{88.80} & 86.57 \\
         & UCF101 & 80.20 & 77.14 & 81.37 & 84.30 & \textbf{85.60} \\
        \rowcolor{green!6} & \textbf{Average} & 76.98 & 72.96 & 78.89 & 80.69 & \textbf{80.96} \\
        \rowcolor{red!5} & \textbf{Time (s)} & 1988 & 5358 & 7060 & 7636 & \textbf{508} \\
        \rowcolor{red!5}\multirow{-14}{*}{8} & \textbf{Mem. (M)} & 4551 & 2888 & 3061 & 3373 & \textbf{528} \\
        \specialrule{.4pt}{0.5pt}{0.5pt}
         & ImgNet & 71.87 & 70.83 & 72.33 & 73.17 & \textbf{73.83} \\
         & Caltech & 95.57 & 95.16 & 96.00 & 96.07 & \textbf{96.77} \\
         & Pets & 91.87 & 93.34 & 92.83 & 93.67 & \textbf{94.00} \\
         & Cars & 83.07 & 71.57 & 83.57 & 83.83 & \textbf{87.43} \\
         & Flowers & 97.07 & 87.84 & 97.00 & 97.60 & \textbf{98.17} \\
         & Food101 & 84.20 & 87.25 & 85.33 & \textbf{87.50} & 87.13 \\
         & Aircraft & 43.40 & 31.21 & 48.40 & 50.83 & \textbf{55.57} \\
         & SUN397 & 74.67 & 72.15 & 75.53 & \textbf{77.23} & 76.80 \\
         & DTD & 69.87 & 63.04 & 71.33 & 72.73 & \textbf{74.07} \\
         & EuroSAT & 84.93 & 73.32 & 92.33 & \textbf{92.43} & 91.57 \\
         & UCF101 & 82.23 & 78.14 & 85.03 & 86.47 & \textbf{87.70} \\
        \rowcolor{green!6} & \textbf{Average} & 79.89 & 74.90 & 81.79 & 82.87 & \textbf{83.91} \\
        \rowcolor{red!5} & \textbf{Time (s)} & 3879 & 10694 & 14070 & 14403 & \textbf{962} \\
        \rowcolor{red!5}\multirow{-14}{*}{16} & \textbf{Mem. (M)} & 4551 & 2888 & 3061 & 3373 & \textbf{528} \\
        \specialrule{1pt}{0.5pt}{2pt}
    \end{tabular}
\end{table*}

Apart from Prompt Tuning (PT), adapter-based methods can be used to transfer large pre-trained vision-language models to downstream tasks.
Table \ref{tab:adapter} compares Skip Tuning with representative adapter-based methods. The details of those comparison methods are listed in Table \ref{tab:adapter}.

\begin{itemize}
    \item LoRA: The default implementation of LoRA \cite{hu2021lora} that adds learnable weights to different blocks of the frozen CLIP model.
    We adopt a learning rate of 1e-3, a batch size of 4, and training epochs of 10.
\item CLIP-adapter. The original implementation of CLIP-adapter \cite{gao2024clipadapter} that attends the learnable adapter module to the frozen CLIP model. The learning rate is 2e-3, the batch size is 32, the number of  training epochs is 100. 
\item CLIP-adapter$^{\dagger}$. A computational efficiency-optimized version of CLIP-adapter \cite{gao2024clipadapter}.
Concretely, the CLIP's last-layer features are cached once before training and subsequently used to update the weights of the adapter module. We adjust the learning rate to 3.5e-3, the batch size to 4, and the number of training epochs to 10. 
\item \textbf{Skip Tuning$^{\dagger}$.} A computational efficiency-optimized version of our Skip Tuning, which sets $\omega=10$ to filter out more shallow features during the LSkip step. 
This variant is used to investigate the performance of Skip Tuning under extreme efficiency constraints.
\end{itemize}




\section{Few-shot Learning Results}
Table \ref{tab:few_shot} presents the few-shot learning performance of different methods across 11 datasets with varying shots of \{1, 2, 4, 8, 16\}.
As can be observed, our Skip Tuning demonstrates superior performance compared to other methods, with significant efficiency advantages in memory usage and time cost across 1-shot to 16-shot settings. 
Additionally, while PromptSRC and MaPLe enhance CoOp's performance, both methods incur increased time cost. 
For instance, PromptSRC\cite{khattak2023promptsrc} improves the average ACC by 2.98\% but increases time cost by $\times$3.7 over the baseline method CoOp.
In contrast, our Skip Tuning achieves both effectiveness and efficiency without compromise, further underscoring the superiority of our approach.


\end{document}

%% file: preamble.tex
\usepackage{booktabs}
\usepackage{multirow}
\usepackage{listings}
\usepackage{tcolorbox}
\usepackage{bbding}
\usepackage{colortbl} 
\usepackage{xcolor}
\usepackage{mdframed}
\usepackage{float}